\documentclass[letterpaper, 10 pt, conference]{ieeeconf}  

\IEEEoverridecommandlockouts                              

\usepackage{graphicx}
\usepackage{amsmath} 
\usepackage{amssymb}  
\usepackage{booktabs}

\title{\LARGE \bf
Stable Object Reorientation using Contact Plane Registration 
}

\author{Richard Li$^{1}$, Carlos Esteves$^{2}$, Ameesh Makadia$^{2}$ and Pulkit Agrawal$^{1}$
\thanks{$^{1}$These authors are with the Improbable AI Lab at
        Massachusetts Institute of Technology, USA and associated with the NSF Institute of Artificial Intelligence and Fundamental Interactions (IAIFI). Correspondence to:
        {\tt\small \{rli14, pulkitag\} @mit.edu}}%
\thanks{$^{2}$These authors are with Google Research, New York.}%
}

\usepackage{algpseudocode} 
\usepackage{textcomp}
\usepackage{gensymb}
\usepackage{algorithm}
\usepackage{hyperref}

\usepackage[dvipsnames,table]{xcolor}
\newcommand{\bgl}{\cellcolor[HTML]{DDDDDD}}
\newcommand{\bgd}{\cellcolor[HTML]{BBBBBB}}

\newcommand{\ourmethod}{CVAE+CMC+MoG5}

\newcommand{\rotation}{\mathbf{R}}
\newcommand{\pointcloud}{\mathcal{X}}
\newcommand{\pointlabel}{\mathcal{Y}}

\renewcommand{\paragraph}[1]{\vspace{.1em}\noindent\textbf{#1}}

\makeatletter
\newcommand{\removelatexerror}{\let\@latex@error\@gobble}
\makeatother

\usepackage{subcaption}

\begin{document}

\let\oldtwocolumn\twocolumn
\renewcommand\twocolumn[1][]{%
    \oldtwocolumn[{#1}{
    \begin{flushleft}
           \includegraphics[width=0.98\textwidth]{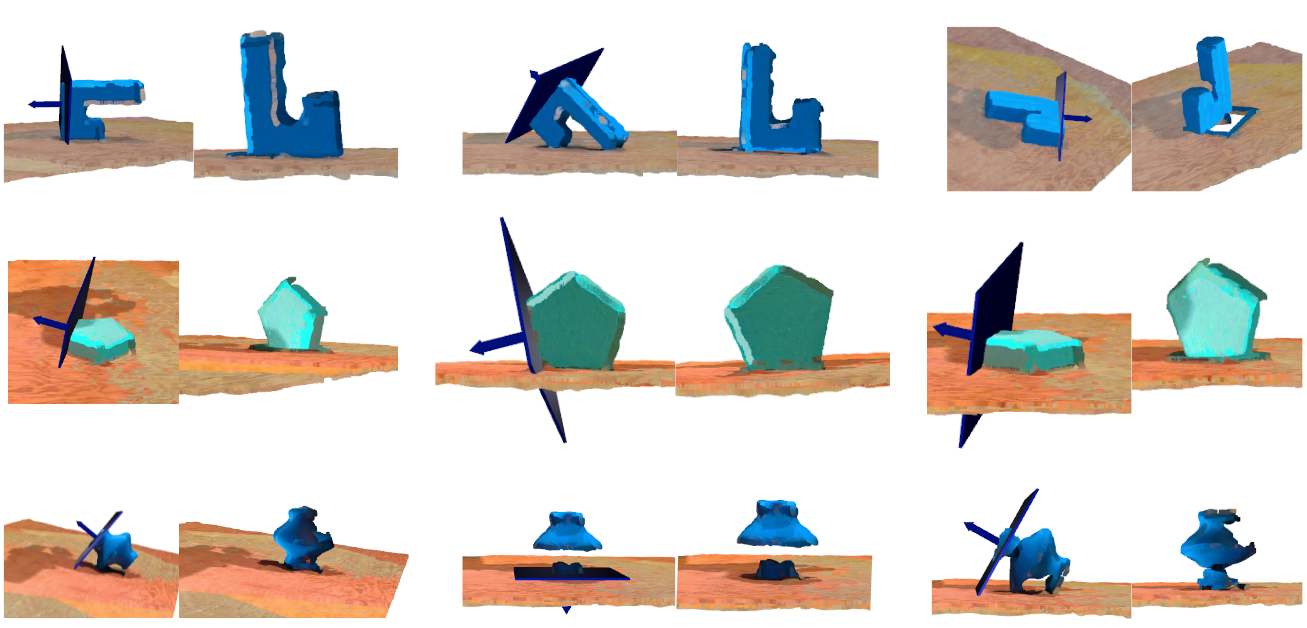}
           
          \captionof{figure}{The image pairs visualize the prediction of our model in reorienting the object in its initial orientation (left) to the height-maximizing stable orientation (right).  The results are shown for three previously \textit{unseen}, \textit{real world} test objects using point clouds captured from the RealSense depth cameras. 
          Our method assumes that objects can be segmented from the scene. The desired transformation is obtained by predicting a contact plane (visualized in blue along with the normal vector) that needs to be aligned with the table to obtain a height-maximizing pose of that object. 
          }
           \label{img:transform_before_after}
    \end{flushleft}
    }]
}

          

\maketitle
\thispagestyle{empty}
\pagestyle{empty}

\begin{abstract}

We present a system for accurately predicting stable orientations for diverse rigid objects. We propose to overcome the critical issue of modelling multimodality in the space of rotations by using a conditional generative model to accurately classify contact surfaces. Our system is capable of operating from noisy and partially-observed pointcloud observations captured by real world depth cameras. Our method substantially outperforms the current state-of-the-art systems on a simulated stacking task requiring highly accurate rotations, and demonstrates strong sim2real zero-shot transfer results across a variety of unseen objects on a real world reorientation task. Project website: \url{https://richardrl.github.io/stable-reorientation/}

\end{abstract}

\section{INTRODUCTION}
\label{sec:intro}
    


Orienting objects is a fundamental operation in many robotic manipulation tasks such as pick-and-place, peg-in-hole, and tool use. Our goal is to develop a system for predicting rotations in the full $\text{SO}(3)$ space for pick-and-place from noisy sensory observations obtained using commodity depth cameras.  We specifically focus on two issues central to rotation prediction: multimodality and generalization to unseen objects from a small number of training shapes. 


The task of fitting a model where similar inputs correspond to very different outputs (with respect to the loss function) is known as the multimodal learning problem. Multimodality can arise from three sources in the rotation prediction task: 1) representational discontinuity, 2) object symmetry, and 3) equivalent optimal rewards.

\begin{enumerate}
    \item Commonly used rotation representations such as Euler angles and quaternions, suffer from discontinuities in the mapping from observation to rotation representation ~\cite{zhou2019continuity,peretroukhin2020smooth,levinson2020svd,mohlin2020probabilistic}. For example, in the 1D angle prediction task, similar observations of an object near the identity pose can have labels at 1 $\degree$ and 359 $\degree$. 
    \item For a symmetric object, multiple rotations of the underlying object model result in the same observation. 
    \item A reward function captures the functional utility of an action. If we consider the rotation predictions as actions given an observation, multiple rotations may correspond to the same optimal reward.
\end{enumerate}







Prior works have addressed the multimodality problem by either implicitly modelling an unnormalized density function \cite{murphy2021implicit, florence2022implicit} or explicitly modelling the distribution of labels using conditional generative models \cite{bishop1994mixture, doersch2016tutorial}.

While multimodality and object generalization may be difficult problems to address in the general case of predicting arbitrary rotations, we consider rotations in a reduced space: namely, the rotations that align one planar contact surface to another. This assumption allows us to introduce specific algorithmic components that can address the two problems. 


To predict the contact plane, we predict the probability that each point is in contact with the target surface when the object is in a correct pose. For example, in this work the correct poses are height-maximizing stable static equilibria. Then, we fit a contact plane to the contact points using random sampling and consensus (RANSAC) plane segmentation  \cite{derpanis2010overview}. 


\begin{figure*}[th!]
      \centering
      \includegraphics[scale=0.4]{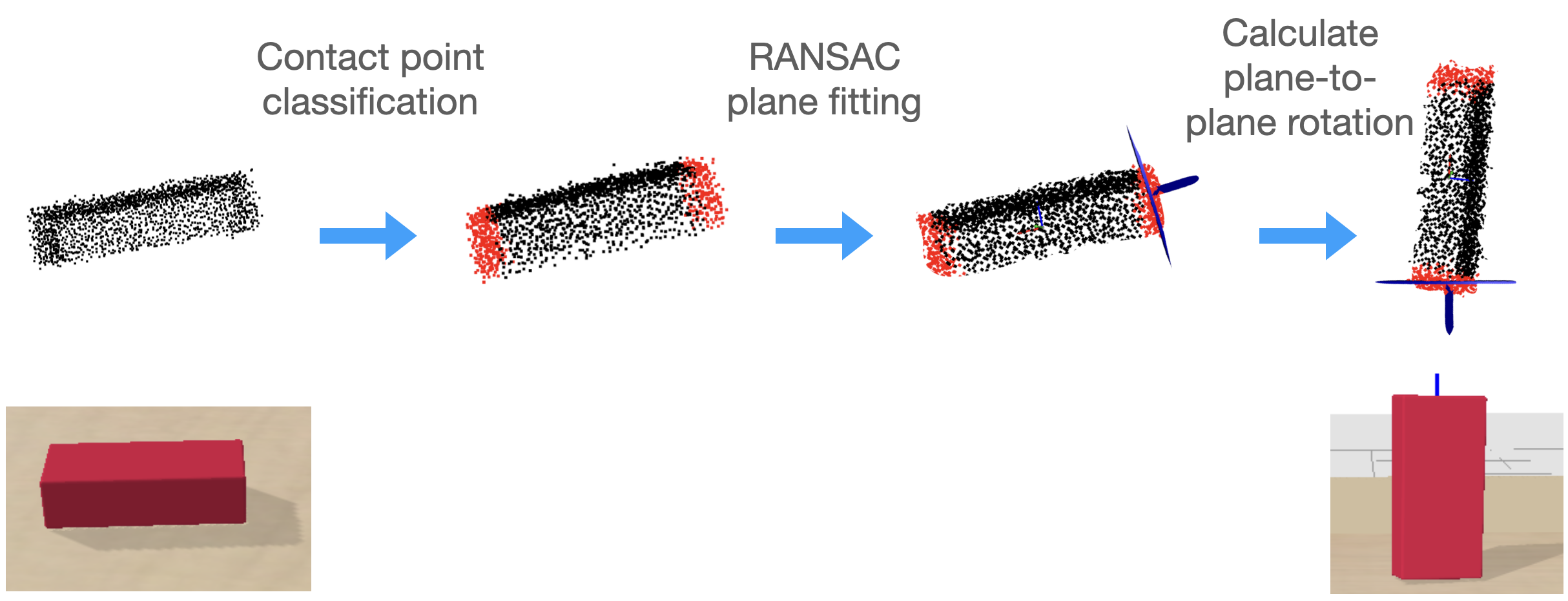}
      \caption{This figure depicts the three steps involved in generating a rotation from an input pointcloud.}
      \label{img:pipeline}
\end{figure*}

Our reformulation of rotation prediction to contact point probability prediction helps with object generalization. The reason object generalization improves is because contact point probabilities are primarily functions of the local geometry. An example of a local feature that our network might learn to determine contact point probability is the discrete Gaussian curvature computed from a point's Euclidean nearest neighbors. Specifically, points on contact planes would be distinguished with curvature 0. Given a new object, as long as the local regions around the contact surface are still similar to those seen in the training set, contact point probability prediction should succeed.



Our contact point probability formulation also helps with producing highly accurate predictions in presence of label multimodality. Our best model uses a CVAE to help with the contact point probability prediction. However, due to VAE-based models' tendency to produce ``blurry" samples, the contact point probabilities will be a weighted average across different binary contact mask labels corresponding to a single input. The CVAE encourages this average to be weighted towards a single contact mask instead of being a uniform average over the contact masks, as would happen with a deterministic model predicting the contact probabilities. Under our formulation, the ``averaged" contact point probabilities can still result in highly accurate predictions, as the RANSAC plane segmentation will find the contact plane with highest inlier support and ignore the outlier points corresponding to the less dense contact region(s).

We evaluate the proposed method on the object reorientation task shown in Figure~\ref{img:transform_before_after}. Given a segmented object point cloud in arbitrary configuration lying on the table, the goal is to predict the rotation that will transform the object into a height-maximizing and stable pose. We show that our method is capable of accurate sim2real transfer across a variety of novel real world objects, and it additionally outperforms state-of-the-art methods on a simulated stacking task that requires highly accurate rotation predictions. 

\section{Method}
\label{sec:method}
\begin{figure}[t!]
      \centering
      \includegraphics[scale=0.22]{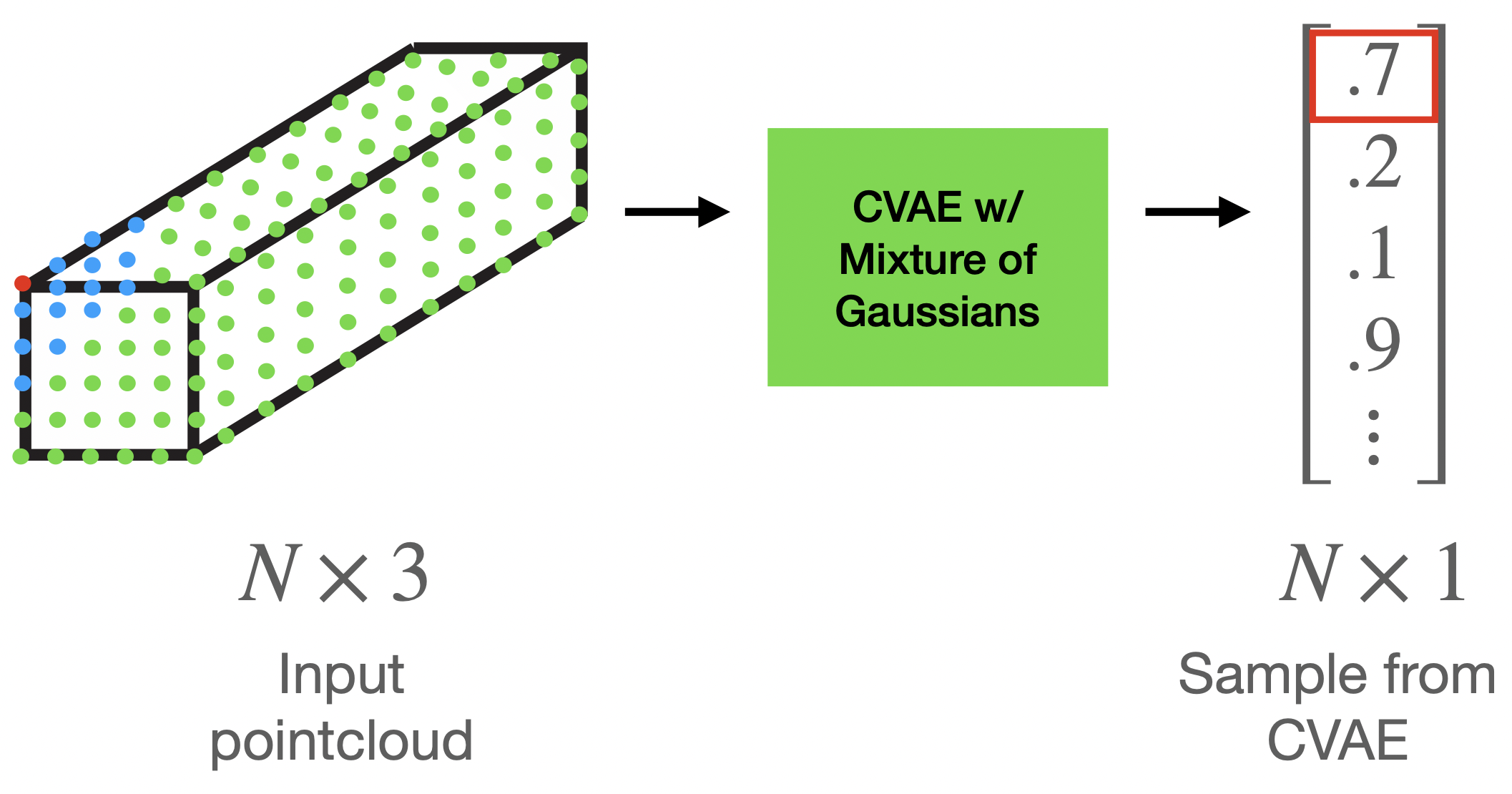}
      \caption{This figure depicts how our CVAE processes the input pointcloud. We highlight in red an input point and the associated predicted contact point probability. We highlight in blue some example points that would be processed by the DGCNN architecture to produce the red contact point probability. The contact point probabilities are thresholded to produce a binary mask.}
      \label{img:cvae_pointcloud}
\end{figure}

Let us assume we are provided a segmented object pointcloud, $\mathcal{X}: \{x_{i} \in \mathbb{R}^3\}_{i=1}^N$ with $N$ points  initialized in an arbitrary orientation on the table. Our goal is to learn a function $f: \mathcal{X} \mapsto \rotation \in \mathbf{SO}(3)$ that outputs a rotation, $\rotation$, that corresponds to placing the object in a height-maximizing and a stable orientation. For this, we first describe our formulation to predict a binary mask corresponding to the object surface that should lie on the table, when the object is in the height-maximizing orientation (see Fig.~\ref{img:cvae_pointcloud} and Sec.~\ref{sec:contact_plane_identification}). 

Next, in Sec.~\ref{sec:contact_plane_registration} we describe a robust procedure for registering the contact plane with the placement surface. The outcome of the registration is the rotation $\rotation$. Finally, we describe neural network details to build the contact classifier in Sec.~\ref{sec:cmc_details}, and real world and simulation results in Sec.~\ref{sec:results}. An overview of the rotation prediction pipeline is in Fig.~\ref{img:pipeline}.


\subsection{Contact Plane Identification}
\label{sec:contact_plane_identification}
We assume that for every pointcloud ($\pointcloud$) in the training set, we have access to a binary ground-truth mask, $\pointlabel: {Y_i \in [0,1]}$, indicating which points on the object are in the contact with the table (i.e., $Y_i = 1$) in the height-maximizing configuration. The procedure for obtaining such labels using minimal manual effort is described in Section~\ref{sec:contact_mask_label_generation}. Using this data we train a classifier for predicting contact points, $c: \mathcal{X} \mapsto \mathcal{Y}$, that takes object-segmented pointcloud ($N \times 3$) as input and outputs per-point probability ($N \times 1$) of a given point being a contact point. We threshold the points with predicted probability $> 0.5$ as the contact points. We now provide details of training this classifier.


\subsubsection{Loss Function} 
We use binary cross entropy (BCE) loss. Because only a small fraction of pointcloud points are contact points, we up-weight the contribution of ground-truth contact points in the BCE loss as following: $$\mathcal{L}(\hat{Y}, Y) = \sum_i -w_{i} [Y_i \cdot \log \hat{Y_i} + (1- Y_i) \cdot \log (1 - \hat{Y_i})] $$
where $\hat{Y}_i, w_i$ is the prediction and the weight of each sample. $w_i$ is defined as:

$$w_i = Y_i \cdot r \cdot \frac{N - \sum_i Y_i}{\sum_i Y_i } + (1- Y_i) \cdot $$

If $r = 1$, the above weighting would equalize the cardinality of positive and negative points with respect to the loss function. Changing $r$ provides expressivity for controlling the precision and recall. We found $r = 2$ to work best. 


\subsection{Rotation Prediction via Contact Plane Registration}
\label{sec:contact_plane_registration}
Given the predicted contact points, we use random sampling and consensus (RANSAC)~\cite{derpanis2010overview} to fit a plane to these points~\cite{zhou2018open3d}. This plane represents the predicted contact surface of the object when it is in in contact with the table in the height-maximizing pose. The choice of RANSAC ensures that plane fitting is robust to incorrect predictions by the classifier.

We find the rotation between the oriented surface normal of the contact plane and the gravity vector (which is the negative surface normal of the table plane) using Rodrigues' formula \cite{moller1999efficiently}.  While we describe our method in the context of orienting the object to be placed on a horizontal surface, it generalizes to placement on any oriented plane whose surface normal is known. 


\subsection{Contact Mask Classifier Details}
\label{sec:cmc_details}

\paragraph{Contact Mask Classifier Baseline (CMC)} This baseline model predicts the probability that each point is a contact point, using a deterministic classification model. When the clusters of contact points are well-separated in Euclidean space, RANSAC plane segmentation can often find a planar fit through one of the clusters.



\paragraph{CVAE + CMC + MoG1} While the baseline model could potentially learn the correct probabilities, we found that combining the baseline model with a generative model improved the training fit and generalization of contact points. We label this model CVAE + CMC + MoG1 to denote its usage of a conditional VAE and a unimodal Gaussian prior. We found that the use of CVAE improved performance and the model more frequently generated samples with distinctive modes, in contrast to the CMC baseline. 



\paragraph{CVAE + CMC + MoG5} 
It is known theoretically that more expressive priors allow VAE models to achieve lower evidence lower bound (ELBO) training objectives \cite{chen2016variational}. In our case, where the space of contact masks has discrete multimodality, it is natural to use a more expressive prior with discrete structure. We modify the CVAE + CMC model by replacing the unimodal Gaussian prior with a Mixture of Gaussians prior with five components. We found that not only did the CVAE + CMC + MoG5 model achieve a lower reconstruction error and KL regularization error, but it also improved accuracy on predicted rotations for unseen objects.




\subsubsection{Data Augmentation} 
\label{sec:data_augmentation}
We apply four types of data augmentation: 1) random rotations in $SO(3)$, 
2) random XY shear, 3) random XYZ scale, and 4) random noise in the pointcloud points sampled from the depth camera noise model (Section~\ref{sec:noisy_pointcloud}). We apply these transformations in a way that ensures correctness of the labelled contact plane. 

\subsubsection{Neural Network Architecture} We used Dynamic Graph CNN (DGCNN) \cite{wang2019dynamic}, a state-of-the-art pointcloud processing architecture with default hyperparameters. 

\subsection{Other Baseline Methods}
\label{sec:other_baselines}
\subsubsection{PCA Oriented Bounding Box Baseline + Ground Truth Face Selection (OBB+GFTS)} Oriented bounding boxes are commonly used as a simple pose representation that is consistent over many different object types. An oriented bounding box consists of eight points that describe the vertices of some cuboid enclosing an object's pointcloud. PCA is commonly used to compute oriented bounding boxes (OBB) for point sets in $\mathbb{R}^n$ \cite{dimitrov2006bounding}. Given an OBB, there are two choices for the height-maximizing contact face. They are the two faces along the longest axis of the OBB. We assume we have an oracle scoring function that tells us which of the two faces will lead to a stable static equilibrium (if one exists), when that face is parallel with the ground plane. 




\subsubsection{Direct Quaternion Prediction Baseline}
We find that direct quaternion prediction, using the representation and ``chordal loss" in \cite{peretroukhin2020smooth} is challenging to optimize with multimodal rotation labels. To ensure that the challenge is caused by multimodality, we ran an ablation where we trained only on five objects in the training set that have no rotational symmetries. This training was successful indicating that multimodality in outputs induced by object symmetry impedes learning. We also experimented with the \textit{ShapeMatchLoss}~\cite{xiang2017posecnn} that was specifically designed to handle rotationally symmetric objects. However similar to~\cite{devgon2020orienting}, we observed that ShapeMatchLoss is hard to optimize. 





\subsubsection{CVAE with Quaternions Baseline}
\label{cvae_quat_baseline}

We would like to understand the importance of modelling rotations with contact regions compared to an explicit representation. Thus, we consider a CVAE formulation whose decoder outputs quaternions. This baseline produces poor results because we are unable to find a good value for the KL-divegence term weight $\beta$  that simultaneously minimizes the reconstruction and KL divergence loss. 




\subsubsection{Implicit PDF Baseline (IPDF)} 
Implicit PDF~\cite{murphy2021implicit} is a state-of-the-art method for learning multimodal distributions over rotation matrices. The main idea is to learn a function $f(x, R)$ that maps an observation $x$ (pointcloud) and rotation $R$ (rotation matrix) to the unnormalized joint density of the input random variables. Then, they acquire a conditional likelihood $p(R|x)$ training objective by approximately integrating this joint density over a discrete set of rotations. 


\section{Experimental Setup}
\label{sec:setup}


\subsection{Object Datasets and Shape Generalization}
\label{sec:object_datasets_and_shape_generalization}
We use objects from two different physical dexterity / stacking games: Bandu and Bausack. These games are played with similar objectives where players adversarially bid tokens (a finite game currency)  to acquire easily stackable pieces and try to force opposing players to stack complicated, unstable pieces. The objects from these games substantially vary in shape and stacking affordances making them a good testbed for evaluating rotation prediction. We create a training, validation, and test set as follows: 15 objects from Bandu for training, 6 objects from Bandu for validation, and 20 objects from Bausack for testing. The train/val split was randomly selected given the constraint that the training set should include a significant number of multimodal objects. We train and validate on only Bandu objects and generalize to the significantly different shapes of the Bausack game.

Five out of fifteen objects in the training set and 11/20 objects in the test set have multimodality in the space of contact mask labels.


\begin{table*}[ht!]
\centering
\caption{Stacking Success Rates and Minimum Angular Errors for Two Object Towers}
\begin{tabular}{llllll}
\toprule
{} &                     OBB+GTFS &                         IPDF &                          CMC &                CVAE+CMC+MoG1 &                CVAE+CMC+MoG5 \\ &&& &&\textbf{(Ours)} \\
\midrule
Barrell              &        0.81 (0.18$\pm$0.23)  &  \bgl 0.97 (0.076$\pm$0.046) &          0.05 (1.4$\pm$0.3)  &         0.9 (0.11$\pm$0.19)  &   \bgd 0.99 (0.053$\pm$0.03) \\
Block                &   \bgd 1.0 (0.051$\pm$0.027) &      0.98 (0.056$\pm$0.028)  &      0.97 (0.043$\pm$0.022)  &  \bgl 0.99 (0.039$\pm$0.021) &  \bgl 0.99 (0.038$\pm$0.022) \\
Bridge               &       0.94 (0.054$\pm$0.03)  &       0.9 (0.058$\pm$0.027)  &      0.92 (0.045$\pm$0.023)  &   \bgl 0.95 (0.04$\pm$0.023) &  \bgd 0.97 (0.034$\pm$0.021) \\
Cone                 &        0.76 (0.42$\pm$0.61)  &   \bgd 1.0 (0.052$\pm$0.026) &  \bgl 0.98 (0.039$\pm$0.024) &   \bgd 1.0 (0.033$\pm$0.019) &   \bgd 1.0 (0.032$\pm$0.019) \\
Cone+Ball            &  \bgd 0.99 (0.043$\pm$0.024) &  \bgl 0.97 (0.054$\pm$0.026) &         0.9 (0.19$\pm$0.63)  &        0.89 (0.15$\pm$0.52)  &  \bgl 0.97 (0.046$\pm$0.027) \\
Cross                &         0.67 (0.6$\pm$0.57)  &        0.69 (0.36$\pm$0.54)  &   \bgd 0.98 (0.065$\pm$0.21) &        0.93 (0.16$\pm$0.39)  &    \bgl 0.94 (0.17$\pm$0.42) \\
Cross Round Top      &        0.73 (0.35$\pm$0.23)  &    \bgl 0.74 (0.37$\pm$0.47) &         0.39 (1.0$\pm$0.71)  &         0.65 (0.56$\pm$0.7)  &    \bgd 0.92 (0.14$\pm$0.36) \\
Cylinder Long        &  \bgd 0.96 (0.021$\pm$0.012) &  \bgl 0.75 (0.052$\pm$0.025) &        0.33 (0.18$\pm$0.29)  &         0.4 (0.13$\pm$0.22)  &        0.38 (0.12$\pm$0.19)  \\
Cylinder Parallelogram &       0.01 (0.48$\pm$0.046)  &       0.86 (0.14$\pm$0.042)  &       0.93 (0.089$\pm$0.22)  &  \bgl 0.96 (0.059$\pm$0.034) &   \bgd 1.0 (0.048$\pm$0.028) \\
Egg                  &    \bgd 0.77 (0.24$\pm$0.31) &        0.67 (0.45$\pm$0.82)  &         0.21 (1.5$\pm$0.28)  &          0.46 (1.0$\pm$1.2)  &    \bgl 0.69 (0.45$\pm$0.71) \\
Foundation           &    \bgd 1.0 (0.051$\pm$0.03) &       0.96 (0.05$\pm$0.029)  &       0.83 (0.06$\pm$0.038)  &   \bgl 0.97 (0.05$\pm$0.028) &      0.93 (0.052$\pm$0.032)  \\
J Block              &       0.33 (0.48$\pm$0.076)  &        0.83 (0.25$\pm$0.39)  &    \bgd 0.89 (0.12$\pm$0.34) &        0.58 (0.66$\pm$0.91)  &    \bgl 0.85 (0.21$\pm$0.48) \\
L Chamfered          &    \bgl 0.97 (0.2$\pm$0.035) &   \bgd 1.0 (0.082$\pm$0.035) &        0.96 (0.11$\pm$0.29)  &        0.96 (0.27$\pm$0.75)  &    \bgd 1.0 (0.084$\pm$0.26) \\
Nut                  &        0.32 (0.36$\pm$0.12)  &         0.4 (0.31$\pm$0.23)  &     \bgl 0.81 (0.25$\pm$0.5) &    \bgd 0.95 (0.12$\pm$0.29) &     \bgl 0.81 (0.3$\pm$0.56) \\
Pentagram Stout      &       0.73 (0.16$\pm$0.089)  &         0.3 (0.33$\pm$0.16)  &         0.18 (1.2$\pm$0.54)  &    \bgl 0.78 (0.44$\pm$0.62) &     \bgd 0.96 (0.3$\pm$0.54) \\
Pyramid w/ Eye        &         0.74 (0.47$\pm$0.2)  &        0.66 (0.55$\pm$0.26)  &  \bgl 0.98 (0.043$\pm$0.026) &   \bgd 1.0 (0.037$\pm$0.023) &   \bgd 1.0 (0.051$\pm$0.033) \\
Rectangular Solid    &  \bgd 0.69 (0.024$\pm$0.012) &  \bgl 0.48 (0.054$\pm$0.029) &        0.37 (0.09$\pm$0.15)  &        0.4 (0.082$\pm$0.15)  &      0.34 (0.078$\pm$0.041)  \\
Rounded Bobble       &    \bgd 0.76 (0.21$\pm$0.27) &     \bgl 0.67 (0.61$\pm$1.0) &          0.01 (1.6$\pm$0.2)  &          0.19 (1.5$\pm$1.2)  &          0.32 (1.0$\pm$1.0)  \\
Vase w/ Hole          &        0.39 (0.86$\pm$0.56)  &    \bgd 1.0 (0.066$\pm$0.03) &        0.57 (0.52$\pm$0.49)  &        0.88 (0.16$\pm$0.27)  &   \bgl 0.94 (0.087$\pm$0.17) \\
Wedge Long           &         0.3 (0.24$\pm$0.29)  &        0.81 (0.098$\pm$0.1)  &   \bgl 0.92 (0.056$\pm$0.15) &    \bgl 0.92 (0.071$\pm$0.2) &   \bgd 0.93 (0.065$\pm$0.16) \\
\bottomrule
\end{tabular}

\label{table:stacking_succ_rate_table}
\end{table*}

\subsection{Training Data Generation and Labelling}
\label{sec:training_data_generation}
We use the PyBullet simulator for our experiments~\cite{coumans2016pybullet}. The starting orientations are generated by randomly dropping objects from a height of 0.4 meters above the table, running forward simulation for 1000 timesteps until the objects have stabilized, then recording their orientation. We use the built-in PyBullet segmentation functionality to get the ground truth object-segmented pointclouds. Random dropping generates variations in starting orientations and occlusion. 

A human expert labels one ``target'' orientation per object. This target orientation corresponds to a height-maximizing and stable pose of the object. However, even with only one target orientation per object, multimodality manifests in the label space of relative rotations due to the object symmetries. The relative rotation between the initial and target orientations is used as the ground truth rotation label.


\subsubsection{Contact Mask Label Generation}
\label{sec:contact_mask_label_generation} 
Given the rotation labels available as quaternions, we follow an automated procedure for generating contact point segmentation masks. A contact mask label is a $N \times 1$ binary tensor that indicates whether each point in the $N \times 3$ pointcloud is a contact point or not. We rotate the observed pointcloud into the stacked position, choose the bottom $k\%$ of points as contact points, and designate the remaining points as non-contact points.


\subsection{Simulating Noisy and Occluded Pointclouds} 
\label{sec:noisy_pointcloud}
To emulate the complexity of real world pointcloud data, we generate training data and evaluate our learned model on a simulated stacking task using realistic noise and occlusion. Our pointcloud noise model is the axial noise model from \cite{ahn2019analysis}, where researchers found that the Intel RealSense D435 depth camera noise was well-approximated by Gaussians with depth-conditional variance, which are centered at the true depth value. We use the same parameters for the noise model as their paper. Our pointclouds are collected by four depth cameras positioned around the table. The segmented pointclouds have occlusions, which are either table-object (object is partially covered by the main flat region of the table or table sides) or, less frequently, object-object (object is partially covered by another object) occlusions.

\section{Results}
\label{sec:results}

We seek to answer the following questions:
\begin{enumerate}
    \item How does the performance of our contact registration method compare with baselines?
    \item What are the effects of various architectural choices on the performance of the proposed system? 
    \item Can our system produce accurate predictions for real world pointclouds?
\end{enumerate}

\subsection{Simulation Two-block Stacking Evaluation}
\label{subseb:stacking_evaluation}
We evaluate the performance of our method and various baselines for rotation prediction by measuring: (a) the accuracy of stacking objects into a tower; (b) and a more granular metric that we call the minimum angular error between normals (\textbf{MAEN}). Definitions of these metrics are:

\paragraph{Stacking Accuracy} is computed using a $score$, which is initially set to 0. Given an ordered list of $K$ objects to be stacked, we increment the $score$ by $+1$ if placing the object increases the height of the tower by the length of the object along the $Z$ axis in the height-maximizing pose. To account for stability the score increment is only made if the object does not fall down and maintains the height-maximizing pose for 500 simulation timesteps after being placed. After all objects are placed, if $score = K$, then the stacking trial is considered to be a success. We measure accuracy over 100 trials.   

\paragraph{MAEN} quantifies the closeness of the predicted contact normal with the set of ground truth contact normals for an object.  We label all optimal contact planes by specifying a list of oriented contact plane normals ($\vec{\eta}_{0:L}$) based on the \textit{mesh} model (not observed pointcloud) of each object. When any of these normals is aligned with the gravity vector, the object enters into a height-maximizing and a stable pose. Let $\theta_l$ denote the rotation angle between the predicted surface normal and the $l^{th} \in [1,L]$ ground-truth normal ($\vec{\eta}_l$). MAEN is defined as $\min_l \theta_l$.

\subsubsection{Quantitative Evaluation} 
\label{sec:results_quantitative}

Our first experiment evaluates rotation prediction for the task of stacking two object towers, where in each trial the bottom object is a cuboid of fixed size and the top object is one of the test objects. We choose a cuboid as the bottom object because it has less surface area than the table and therefore stable placement of the object requires higher accuracy in rotation prediction. For each test object, accuracy evaluated  over 100 trials is reported in Table \ref{table:stacking_succ_rate_table}. The mean and standard deviation of the MAEN are included in the parentheses in Table \ref{table:stacking_succ_rate_table}.

\begin{figure}[t!]
      \centering
      \includegraphics[scale=0.5]{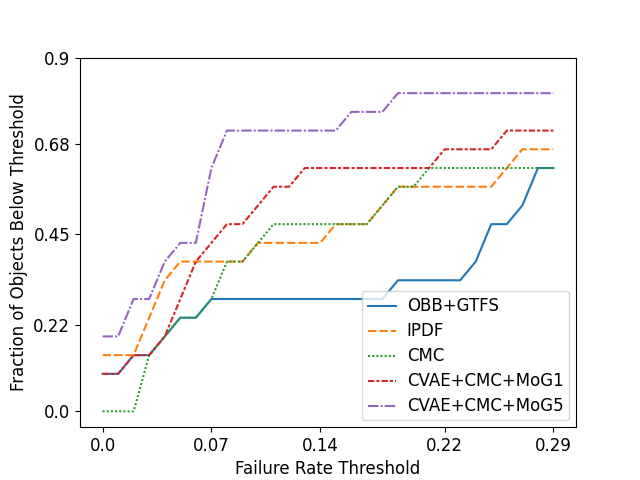}
      \caption{Curve showing (y-axis) fraction of objects types below a certain failure rate, associated with each model, (x-axis) for various failure rates (best viewed in color).}
      \label{img:auc_curve}
\end{figure}


The results reveal that our method, \ourmethod{}, outperforms all the baselines. Further quantification of which model performs best across all object types is presented in Fig. \ref{img:auc_curve}. The figure plots the fraction: $\frac{\text{\# objects whose failure rate} \leq \text{failure rate threshold}}{\text{total \# objects}}$ on the y-axis and incremental failure rate thresholds on the x-axis. \footnote{The object failure rate is is $1-\tt{success\_rate}$ (Table \ref{table:stacking_succ_rate_table}).} From this plot, we can see that for any failure threshold we select, our \ourmethod{} model's failure rates are below the threshold, for the highest number of objects, compared to other models. Our method stays under a $19\%$ failure rate threshold for 16/20 objects, while the second best stays under that threshold for only 12/20 objects. 

\begin{figure*}[th!]
      \centering
      \includegraphics[scale=0.3]{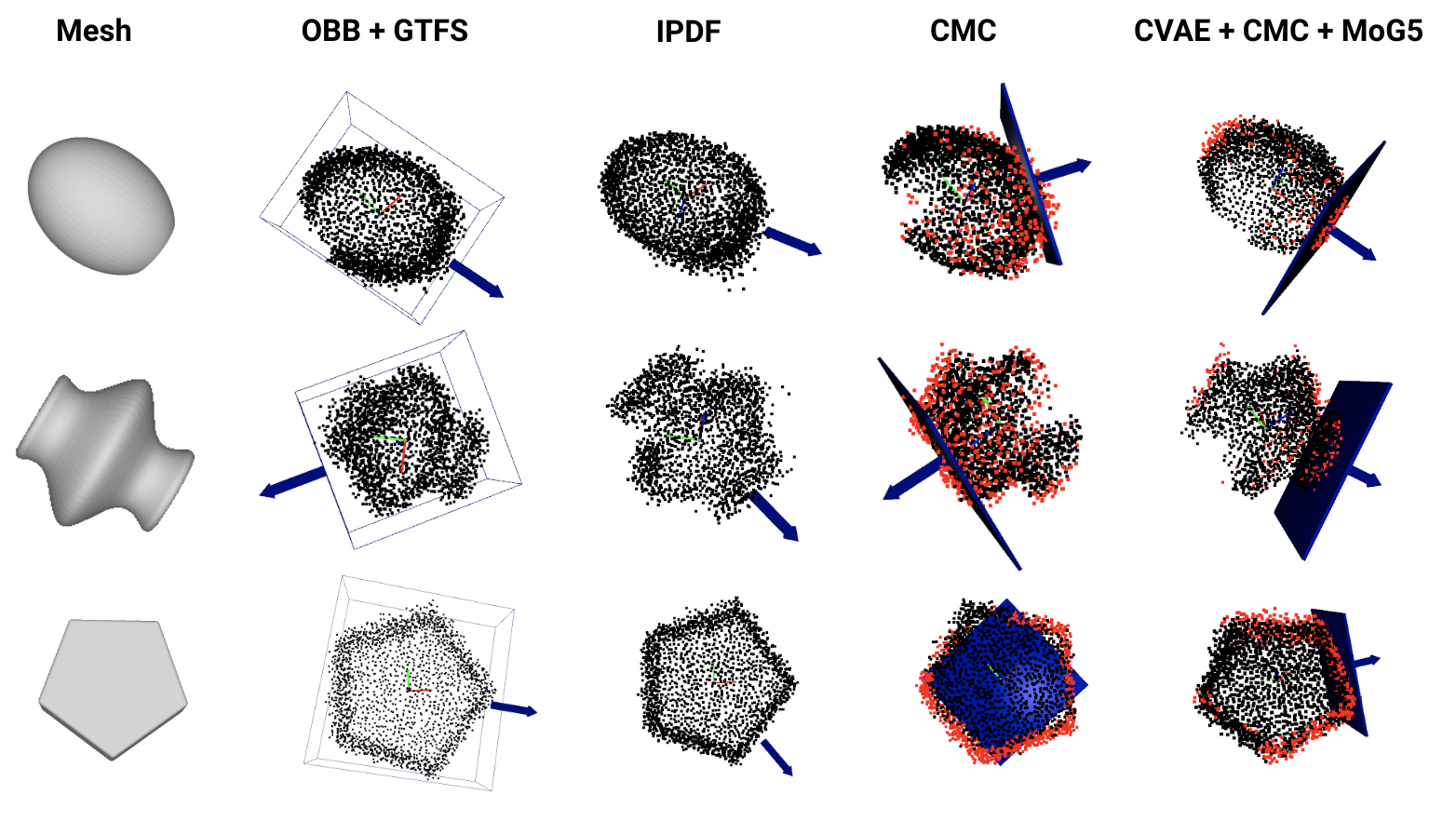}
      \caption{(Left to right: model name) Mesh, OBB + GTFS, IPDF, CMC, CVAE + CMC + MoG5. (Top to bottom: object name) Egg, Vase w/ Hole, Pentagram Stout. Although IPDF predicts rotations and does not directly generate normal vectors, we visualize here the gravity vector that has been rotated by the \textit{inverse} of the rotation predicted by IPDF, to get a similar gauge on how well the IPDF predicted rotation is aligning the correct contact plane normal to the gravity vector.}
      \label{img:normals_comparison}
\end{figure*}

\subsubsection{Simulation Qualitative Analysis} 

When the CMC method fails, it often produces large, arbitrary swathes of contact points that do not correspond to contact regions (Fig. \ref{img:normals_comparison}). 
In comparison, \ourmethod{} produces thinner sets of contact points, which lead to more accurate contact plane fitting and higher accuracy. Both models tend to incorrectly predict multiple modes per sample for multimodal objects, and correctly predict one mode for unimodal objects. For the CVAE, this behavior of multiple mode prediction is likely caused by upweighting the $r$ term in the loss function, which increases the rate of false positives. Since the RANSAC plane segmentation can exclude false positives at test time, we find that false positives are preferable to having many false negatives and too little contact points.

\subsection{Sim2Real Real World Results}
\begin{table}[ht!]
\centering
\caption{Tabletop Reorientation Success Rates}
\begin{tabular}{llllll}
\toprule
{} &                     Simulation &                         Real world  &        \\
\midrule
Barrell              &       5/5  &  5/5 \\
Bridge               &       5/5  &       5/5  \\
Cone+Ball            &   5/5 &  5/5 \\
Cross                &        5/5  &        3/5  \\
J Block              &       5/5  &        5/5  \\
Nut                  &       4/5  &         5/5  \\
Pentagram Stout      &       5/5  &         5/5  \\
Pyramid w/ Eye        &        5/5  &        5/5  \\
Vase w/ Hole          &        5/5  &    5/5 \\
\bottomrule
\end{tabular}

\label{table:real_world_table}
\end{table}

In this section, we characterize the performance of our trained model on 3D-printed, real world versions of the unseen test objects. Our evaluation metric is the success rate. Success is determined by a human evaluator as to whether the predicted contact plane closely (within 15 degrees) matches one of the height-maximizing contact planes of the true object. Because we are forced to use a relatively small number of trials in real world experiments, we match the sim and real world pose distributions by  manually setting the object poses in pyBullet to the real world object poses. Additionally, we ensure our real world/simulation orientation distribution is realistic by approximately uniformly sampling within the various contact modes/stable static equilibria modes of the object. We do this because in the wild, objects are often set by humans in stable static equilibria, and the distribution of these human-set equilibria may be significantly different from the distribution of static equilibria found by dropping the objects from the sky (which is the method we used to generate our training data in sim).

We show qualitative trials of our real world orientation predictions in Fig. \ref{img:transform_before_after}. We see that these transformations are very accurate at reorienting the objects into an upright pose.




\subsubsection{Analysis} We find that in nearly all objects, the real world Sim2Real transfer succeeds 5/5 times. There were two important details we implemented to get such results. The first is using the ``High Accuracy" preset for the RealSense 415 cameras, which removes noise at the cost of also removing some points on the actual object itself. The second detail is appropriately scaling the real world pointclouds. Because we did not carefully calibrate the poses and intrinsic parameters of the cameras in the real world to match those in simulation, the real world pointclouds for a few objects ended up being much smaller.

\section{Related work}
\label{sec:related_work}
 
\paragraph{Multimodal Modelling}
Substantial progress has been made in modelling multimodal distributions using conditional generative models such as conditional variational autoencoders (CVAE)~\cite{doersch2016tutorial}. 
However, CVAEs still have difficulty producing crisp samples across all the data modes, in part due to difficulty optimizing the competing objectives of accurate reconstruction and KL divergence minimization ~\cite{higgins2016beta}. 
 Another way of modelling multimodal distributions that has worked well in robotics is classification into discretized bins of continuous quantities (such as rotations) ~\cite{pinto2015supersizing,agrawal2016learning,nair2017combining,mohlin2020probabilistic, mahendran2018mixed}. This discretization can be challenging because if the bins are too coarse then resulting predictions will be inaccurate. Alternatively, if many small bins are constructed, it reduces the capability for information sharing across samples. Finally, a current state-of-the-art method is implicitly modelling the unnormalized density of (observation, rotation) pairs \cite{murphy2021implicit}. This method converts multimodal rotation prediction to unimodal density prediction. 



\paragraph{Block Stacking}
While much prior work has investigated stacking towers of simple shapes, such as cuboids or triangular prisms, \cite{deisenroth2011learning, duan2017one, nair2018overcoming, janner2019reasoning, openai2021asymmetric, li2020towards}, less work has demonstrated stacking complex, irregular 3D shapes. We describe works that do operate on complex shapes. \cite{furrer2017autonomous} showed that a mesh reconstruction pipeline could be used to load natural limestones into a simulator and find stable poses for stacking. Errors in reconstruction may seriously affect downstream control based on the mesh.



 \cite{kroemer2018kernel} uses a learned, kernel-based scoring function to score translations of the pointcloud of the object-to-be-stacked. They assume the objects start off in the correct orientation, while our focus is on learning how to orient the object. \cite{wang2019dynamic} used a combination of learning from demonstration and reinforcement learning to learn to stack objects. Compared to their work, we consider more complex objects requiring large out-of-plane rotations and focus on introducing a new rotation representation to handle this complexity.

\paragraph{Pose Estimation for Symmetric Objects}
Many of the objects we consider in this work have discrete or continuous symmetries that introduce multimodality in the label space of rotations. Past works have investigated learning implicit function representations \cite{murphy2021implicit}, reducing symmetry groups to canonical poses \cite{pitteri2019object}, and matching an input image to synthetically rendered views, in order to retrieve poses for symmetric objects \cite{sundermeyer2020augmented}. 


\section{Acknowledgements} We would like to thank Branden Romero for helping 3D print the physical objects, and Abhishek Gupta and Tao Chen for feedback on the writing. We also thank the NSF Institute of Artificial Intelligence and Fundamental interactions (IAIFI) and Sony for their financial support.

\bibliographystyle{IEEEtran}
\bibliography{IEEEabrv,IEEEicra2022}{}

\end{document}